%% file: main.tex
\documentclass[10pt,twocolumn,letterpaper]{article}

\usepackage[pagenumbers]{iccv} 


\definecolor{iccvblue}{rgb}{0.21,0.49,0.74}
\usepackage[pagebackref,breaklinks,colorlinks,allcolors=iccvblue]{hyperref}
\usepackage{bbm}
\usepackage{algorithm2e}
\usepackage{multirow}

\SetKwComment{Comment}{// }{}
\RestyleAlgo{ruled}

\newcommand{\argmax}{\mathop{\mathrm{argmax}}\limits}
\newcommand{\methodName}{\textsf{LoGIC}}
\newcommand{\speaker}{\textsf{Speaker}}
\newcommand{\listener}{\textsf{Listener}}

\usepackage{xcolor}
\usepackage{enumitem}

\def\paperID{12722} 
\def\confName{ICCV}
\def\confYear{2025}

\title{Emergent Natural Language with Communication Games for Improving Image Captioning Capabilities without Additional Data}

\author{Parag Dutta and Ambedkar Dukkipati\\
Department of Computer Science and Automation\\
Indian Institute of Science, Bangalore, KA, IN - 560012\\
{\tt\small \{paragdutta, ambedkar\}@iisc.ac.in}
}

\begin{document}

\maketitle

\begin{abstract}
Image captioning is an important problem in developing various AI systems, and these tasks require 
large volumes of annotated images to train the models. Since all existing labelled datasets are already used for training the large Vision Language Models (VLMs), it becomes challenging to improve the performance of the same. Considering this, it is essential to consider the unsupervised image captioning performance, which remains relatively under-explored. To that end, we propose \methodName~(\underline{L}ewis C\underline{o}mmunication \underline{G}ame for \underline{I}mage \underline{C}aptioning), a Multi-agent Reinforcement Learning game. 
The proposed method consists of two agents, a \speaker~and a \listener, with the objective of learning a strategy for communicating in natural language. 
We train agents in the cooperative common-reward setting using the GRPO algorithm and show that improvement in image captioning performance emerges as a consequence of the agents learning to play the game. 
We show that using pre-trained VLMs as the \speaker and Large Language Model (LLM) for language understanding in the \listener, we achieved a $46$ BLEU score after fine-tuning using \methodName without additional labels, a $2$ units advantage in absolute metrics compared to the $44$ BLEU score of the vanilla VLM. Additionally, we replace the VLM from the \speaker with lightweight components: (i) a ViT for image perception and (ii) a GPT2 language generation, and train them from scratch using \methodName, obtaining a $31$ BLEU score in the unsupervised setting, a $10$ points advantage over existing unsupervised image-captioning methods.
\end{abstract}

\section{Introduction}
\label{Section:Introduction}
Generating natural language descriptions for images has gained attention as it involves making machines learn how humans perceive, understand, and interpret the world. The research on image captioning has made significant progress in the past decade~\cite{DeepLearningApproachesonImageCaptioning-Areview,ShowAndTell-LessongsLearnedFromThe2015MSCOCOimageCptioningChallenge}. Most of the proposed methods, however, learn a deep neural network model to generate captions (descriptions) given an input image in the supervised setting, \ie from a dataset of manually annotated image-caption pairs. However, the acquisition of these paired data is a costly and labor-intensive process. The diversity of image and their corresponding captions in the existing image captioning datasets, such as Microsoft COCO~\cite{10.1007/978-3-319-10602-1_48} are limited to around $100$ objects which poses a significant challenge due to the generalization gap when deploying these systems in the real world.

For instance, a straightforward way to implement image retrieval using natural language queries on mobile devices is to pre-compute the image captions and rank the images based on the similarity of the captions with the user query. However, a domain gap between the training images and the user-captured images exists. Fine-tuning on the new domain is difficult since users are not keen on sharing their private images with the software developers, and even then, manually annotating these images will incur additional costs. This is an open problem for many tech companies, such as Apple, Samsung, and Google, that wish to incorporate natural language gallery search into their AI software suite.


Consequently, some recent works have approached the problem from the direction of unsupervised image captioning~\cite{feng2019unsupervisedimagecaptioning, Laina_2019_ICCV, YANG2023102490, ijcai2020p128, CAO2020419}. These works broadly use the following two techniques: (i) Adversarial representation learning and (ii) Object Detection, with each of them introducing its own set of challenges. For instance, adversarial approaches can have mode collapse due to the saddle point optimization objective and require significant computational overhead trying to prevent the same, \eg reconstruction and consistency losses. Again, the object detector is limited by the diversity of the objects that the model has already seen in the training set. As it so happens, Microsoft COCO also has object detection annotations, and therefore, detection algorithms trained on this dataset often work well for the captioning task. When shown out-of-distribution images, however, these approaches fail to detect or incorrectly detect objects in the scene, leading to low-quality captions.

We, on the other hand, investigate the problem from the angle of how human intelligence works. For instance, when children notice a novel object, rather than `hallucinating', they describe the object to their parents by composing a description using a combination of `primitives' that they already know about. Then, given the feedback from the parents, they learn about novel objects. To this end, we propose the use of a Multi-agent Reinforcement Learning (MARL) based framework to solve the problem of Image Captioning, which is more aligned to how humans learn.

Lewis signaling games are a class of simple communication games for simulating the emergence of language. In these games, two agents must agree on a communication protocol in order to solve a cooperative task. In their original form, Lewis signaling games involve two agents: a \speaker~and a \listener. The \speaker~observes a random state from its environment, \eg an image, and sends a signal to the \listener. The \listener~then undertakes an action based on this signal. Finally, both agents are equally rewarded based on the outcome of the \listener~action. The resolution of this cooperative two-player game requires the emergence of a shared protocol between the agents~\cite{lewis2002communication,crawford1982strategic}.

We extend the Lewis game~\cite{Lewis} by constraining the signals to natural language and propose \underline{L}ewis C\underline{o}mmunication \underline{G}ame for \underline{I}mage \underline{C}aptioning (\methodName). In \methodName, the \speaker~is shown an image, and the \listener~has to identify the same image that the \speaker~was shown amongst a set of $K$ images in which $K-1$ images are `distractors'. By scaling up the number of distractors to the order of thousands and using a Vision Language Model~\cite{Qwen-VL, lu2024deepseekvl, liu2023visualinstructiontuning} in the \speaker and Large Language Models (LLMs)~\cite{brown2020languagemodelsfewshotlearners, touvron2023llamaopenefficientfoundation, gunasekar2023textbooksneed} for language understanding in the \listener, we learn a co-operative strategy to solve \methodName~. 
However, since these models are very large and have large inference time, we replace the \speaker with smaller models: Vision Transformers~\cite{dosovitskiy2021imageworth16x16words, bao2022beitbertpretrainingimage} for image understanding, Language Models~\cite{devlin-etal-2019-bert, liu2019robertarobustlyoptimizedbert, yang2020xlnetgeneralizedautoregressivepretraining} (specifically GPT2~\cite{radford2019language}) for caption generation and find that image captioning is an emergent behavior of the learned strategy required to solve \methodName. Although the training computation requirements are high, it must be noted that the inference (caption generation) on the smaller models can be done on almost any modern computer, such as a gaming laptop with a $4$GB GPU with CUDA support or an Apple silicon Macbook with Metal Performance Shaders (MPS) support.

\noindent
\textbf{Contributions.}
\begin{itemize}
    \item We formulate the task of unsupervised image captioning as a multi-agent reinforcement learning game.
    \item We propose a policy gradient-based algorithm to learn a strategy to play the game.
    \item We provide experimental results and show the learning to play \methodName~can achieve state-of-the-art results in the image captioning task.
    \item We ground our proposed algorithm by formulating it with reference to existing representation learning techniques.
\end{itemize}

\section{Related Works}
\label{Section:Related}
Supervised image captioning has been studied extensively in the literature and most methods use an image encoder (CNN/ViT based) and then decode the encoded representation using a language decoder (RNN/Transformer based) to generate a sentence describing the image~\cite{Fidler_2013_CVPR, ShowAndTell-LessongsLearnedFromThe2015MSCOCOimageCptioningChallenge, 10.1007/978-3-642-15561-1_2, li-etal-2011-composing, chen2014learningrecurrentvisualrepresentation, 7298878, 7298754, kiros2014unifyingvisualsemanticembeddingsmultimodal, mao2014explainimagesmultimodalrecurrent, karpathy2015deepvisualsemanticalignmentsgenerating}. These models are trained to maximize the probability of generating the ground truth caption conditioned on the input image and optionally additional multi-modal information. It has been shown that token prediction can be learned using policy gradient-based methods, albeit in the supervised setting~\cite{ranzato2016sequenceleveltrainingrecurrent}. 

Since paired image caption datasets are expensive to obtain, some studies have been conducted to explore the possibility of reducing the paired information needed, \ie image captioning in the semi-supervised setting~\cite{anderson2018partiallysupervisedimagecaptioning, jin2023selfsupervisedimagecaptioningclip, chen2017showadapttelladversarial}, or utilizing unpaired dateset of images and sentences~\cite{feng2019unsupervisedimagecaptioning, Laina_2019_ICCV,gu2019unpairedimagecaptioningscene, ijcai2020p128, honda-etal-2021-removing,8970902, meng2022objectcentricunsupervisedimagecaptioning}. \citet{gu2019unpairedimagecaptioningscene} explored learning simultaneously from multiple data sources with auxiliary objectives to describe a variety of objects unseen in paired image-caption datasets. Most of these works assume that a pre-trained object detector is available for extracting visual concepts from an image, which acts as the link between the visual and the language domains. Some of them also utilize adversarial training to align the visual and language domains. In the partial supervision setting, ~\cite{10.1007/978-3-030-01267-0_21} comes somewhat close to our approach, however, they used supervision for alignment and to prevent mode-collapse.

Recently, some works~\cite{feng2019unsupervisedimagecaptioning, Laina_2019_ICCV, YANG2023102490, ijcai2020p128, CAO2020419} have been looking into fully unsupervised image captioning. Many of them use the aforementioned techniques, such as object detectors and adversarial training, for aligning the visual and language domains. Many of them also exploit the language domain, which is rich in visual concepts. Supervision comes through the image-recognition models, which are used to detect objects in the images.

Scaling up the parameters in Vision-Language Models (VLMs)~\cite{liu2023visualinstructiontuning, Qwen-VL, lu2024deepseekvl} has also been successful for image captioning tasks. Recent advances in the technologies can be found in Bordes \etal~\cite{bordes2024introductionvisionlanguagemodeling}. We specifically use the models categorized as `Pre-trained backbones' as those have clearly separable vision encoders and language generators.

Lewis signaling games~\cite{Lewis, lewis2002communication} were proposed for learning a convention (protocol) among agents in the cooperative setting. Subsequently, it has been extensively used to study `emergent' language in multi-agent RL~\cite{ueda2024lewisssignalinggamebetavae, mu2022emergentcommunicationgeneralizations, emcom_scale, NEURIPS2022_093b08a7} in order to gain insights into the evolution of natural languages in human society. To the best of our knowledge, ours is the first work to design natural language communication in Lewis games at scale and realize that image captioning is an emergent behavior learned by the optimal cooperative strategy to play the game. We use the Group Relative Policy Optimization~\cite{shao2024deepseekmathpushinglimitsmathematical} algorithm for backpropagating the RL cost to the partially differentiable entities.

\section{Methodology}
\label{Section:Methodology}

\subsection{Background on RL}
We model a sequential decision-making problem under uncertainty as a Markov Decision Process (MDP) in the finite horizon discounted reward setting specified by the tuple $( \mathcal{S}, \mathcal{A}, \mathcal{T}, r, \rho, \gamma)$, where $\mathcal{S}$ is the set of possible states of an environment, $\mathcal{A}$ is the set of actions that can be taken by the agent, $\mathcal{T} : \mathcal{S} \times \mathcal{A} \rightarrow \Delta \mathcal{S}$ is the transition function, $r: \mathcal{S} \times \mathcal{A} \rightarrow \mathbb{R}$ is the reward function, $\rho$ is the initial state distribution and $\gamma \in [0, 1)$ is the discount factor.



While there is only one agent in canonical RL, multi-agent RL deals with sequential decision-making problems with more than one agent. One can also formalize this using game-theoretic models. One such model is a normal-form game (or a strategic-form game), which consists of a finite set of agents $\mathcal{I}={1,...,n}$, where $n\geq2$. For each agent $i\in\mathcal{I}$, $\mathcal{A}_i$ denotes corresponding action set and reward function $\mathcal{R}_i:\mathcal{A}\rightarrow\mathbb{R}$, where $\mathcal{A}=\mathcal{A}_1\times \ldots\times\mathcal{A}_n$. 
A common-reward game is a normal-form game where all agents receive the same reward, \ie $\mathcal{R}_i=\mathcal{R}_j$, for all $i,j \in \mathcal{I}$.

\begin{algorithm}[t]
\KwIn{$\mathcal{D}$: Dataset of images
}

\KwOut{$\theta^*$: optimal parameters for co-operative strategy to play \methodName}
\textbf{Initialization:} $\mathcal{M}_1, \mathcal{M}_2$ with joint parameters $\theta$\;
\While{not converged}{
    \Comment{Sample images from dataset}
    $s,{s'}_1,{s'}_2,...,{s'}_{K-1}\sim\mathcal{D}$\;
    \Comment{Generate message (caption)}
    $m\sim\pi_1(\mathcal{M}_1^\mathrm{Vision Encoder}(s))$\;
    \Comment{Get image representations}
    $v_1,...,v_K=\mathcal{M}_2^\mathrm{Vision Encoder}(s,{s'}_1,...,{s'}_{K-1})$\;
    \Comment{Get message representation}
    $v_m=\mathcal{M}_2^\mathrm{Language Encoder}(m)$\;
    \Comment{Compute probabilities}
    $\mathrm{Prob}[v_k=s]=\frac{\exp(<v_m\cdot v_k>)}{\sum_{j=1}^K\exp{<v_m\cdot v_j>}}$\;
    \Comment{Backprop losses and update}
    $\theta^+\leftarrow\theta-\eta\nabla_\theta(\mathcal{L}_\mathrm{Speaker}+\lambda\mathcal{L}_\mathrm{Listener})$
}
\caption{Algorithm to solve \methodName}
\label{alg:algorithm_active}
\end{algorithm}

\begin{table*}[t]
    \centering
    \begin{tabular}{|l|c|c|c|c|c|c|c|c|}
        \hline
        Method & BLEU-1 & BLEU-2 & BLEU-3 & BLEU-4 & METEOR & ROUGE & CIDEr & SPICE \\
        \hline
        UIC~\cite{feng2019unsupervisedimagecaptioning} & 41.0 & 22.5 & 11.2 & 5.6 & 12.4 & 28.7 & 28.6 & 8.1 \\
        \citet{Laina_2019_ICCV} & - & - & - & 6.5 & 12.9 & 35.1 & 22.7 & - \\
        IGGAN~\cite{10609416} & 42.2 & 23.1 & 11.8 & 6.5 & 13.1 & 30.5 & 28.7 & 8.2 \\
        \citet{YANG2023102490} & 43.3 & 24.2 & 11.7 & 6.1 & 12.5 & 30.2 & 25.9 & 8.4 \\
        ViT+GPT (unsup) & \textbf{63.4} & \textbf{39.5} & \textbf{24.0} & \textbf{16.3} & \textbf{19.7} & \textbf{44.9} & \textbf{62.4} & \textbf{12.8} \\
        \hline
        Qwen VL & 70.4 & 63.0 & 38.4 & 27.8 & 24.6 & 51.1 & 82.2 & 16.9 \\
        Qwen VL + \methodName & \textbf{72.9} & \textbf{65.4} & \textbf{40.1} & \textbf{29.0} & \textbf{25.5} & \textbf{51.8} & \textbf{83.0} & \textbf{17.3} \\
        \hline
    \end{tabular}
    \caption{The results for comparison of our method with existing unsupervised image captioning methods as baselines. For all metrics, the higher, the better. We can see that our method is quite good when compared to existing unsupervised approaches, and can also improve the performance of pre-trained VLMs without additional labelled data.}
    \label{Table: Main Results}
\end{table*}

\subsection{Natural Language Lewis Game}

The Lewis signaling game can be formalized as a two-player cooperative common-reward normal-form game. We extend the idea by constraining the language generated by the \speaker~to the domain of natural languages.

\methodName~consists of 2 agents where $i=1$ is the \speaker~and $i=2$ is the \listener.
\begin{enumerate}
    \item \speaker: The state space of the \speaker~is the domain of RGB images, \ie $\mathcal{S}_1=\mathbb{R}^{h\times w\times 3}$ where $h$ and $w$ are some fixed height and width of the images, for instance, $s\in\mathcal{S}_1$ is one such image. The action space of the \speaker~is $\mathcal{A}_1\subseteq|\mathcal{V}|^{T_{max}}$ where $\mathcal{V}$ is the vocabulary of the Speaker, and $T_{max}$ is the maximum permissible number of words (or tokens) the \speaker~is allowed to communicate. Thus a typical message is of the form $m=\{w_1,w_2,...,w_{T}\}$ where $w_t$s are the words sampled from the vocabulary under some \speaker~policy $\pi_1(s)$ (conditioned on the state) and $T<=T_{max}$. Sampling the message is a sequential decision-making process since the current word has an effect on the meaning of the sentence in the future and thus on the outcome received. Hence, the \speaker~must be equipped with an RL policy $\pi_1$ to sample the message.
    \item \listener: The \listener~accepts the message $m$ as input, thus the state space $\mathcal{M}_2\subseteq|\mathcal{V}|^T$. However, the notable distinction in this case is that the message $m$ need not be parsed sequentially by the \listener. The image shown to the \speaker~(state $s$) is mixed in with $K-1$ distractor images. The action space of the \listener~$\mathcal{A}_2=\{1,2,...,K\}$, \ie the action of the \listener~$a\in[K]$. Sampling $a$ is a one-step decision, and hence the policy $\pi_2(m)$ can be modeled as a contextual Multi-armed Bandit (which is a special case of MDPs where the next state is not dependent on the current decision).
\end{enumerate}

The common reward obtained by the agents is $1$ if the \listener~chooses $k$ and $s$ is indeed the $k^{th}$ image among the distractors, otherwise both of the agents receive $0$ reward, \ie $\mathcal{R}_1=\mathcal{R}_2=\mathbbm{1}\{a=k\}$.

\subsection{A Practical Algorithm for \methodName}
We use $\mathcal{M}_1$ and $\mathcal{M}_2$ to denote the \speaker~and \listener~models respectively. $\mathcal{M}_1$ and $\mathcal{M}_2$ have modules called encoders and decoders. For instance, the encoder in $\mathcal{M}_1$ and Vision Encoder in $\mathcal{M}_2$ may be modeled using a Vision Transformer~\cite{dosovitskiy2021imageworth16x16words}, the decoder in $\mathcal{M}_1$ and the language encoder in $\mathcal{M}_2$ using a Transformer~\cite{transformer}, and the decoder in $\mathcal{M}_2$ using a Multi-layer Perceptron (MLP).

\paragraph{Reward shaping}
First of all, we relax the reward function of the Lewis game $\mathcal{R}_i=\mathbbm{1}\{a=k\}$ and define the new reward function to be $\mathcal{R}_i=\mathbb{P}[a=k]$, where the probability is realized using the parameters of the decoder in $\mathcal{M}_2$, \ie the reward is $\pi_2(a=k|m)$, the probability assigned by the decoder in $\mathcal{M}_2$ that $s$ is the image corresponding to the description message $m$. Since the objective of the cooperative game is to maximize the expected reward, we can see that the new reward function preserves the original goal:
\begin{align*}
    \max \mathbb{E}[\mathcal{R}_i]=\max \mathbb{E}[\mathbbm{1}\{a=k\}]=\max \mathbb{P}[a=k]
\end{align*}

\begin{figure*}
    \centering
    \includegraphics[width=\linewidth]{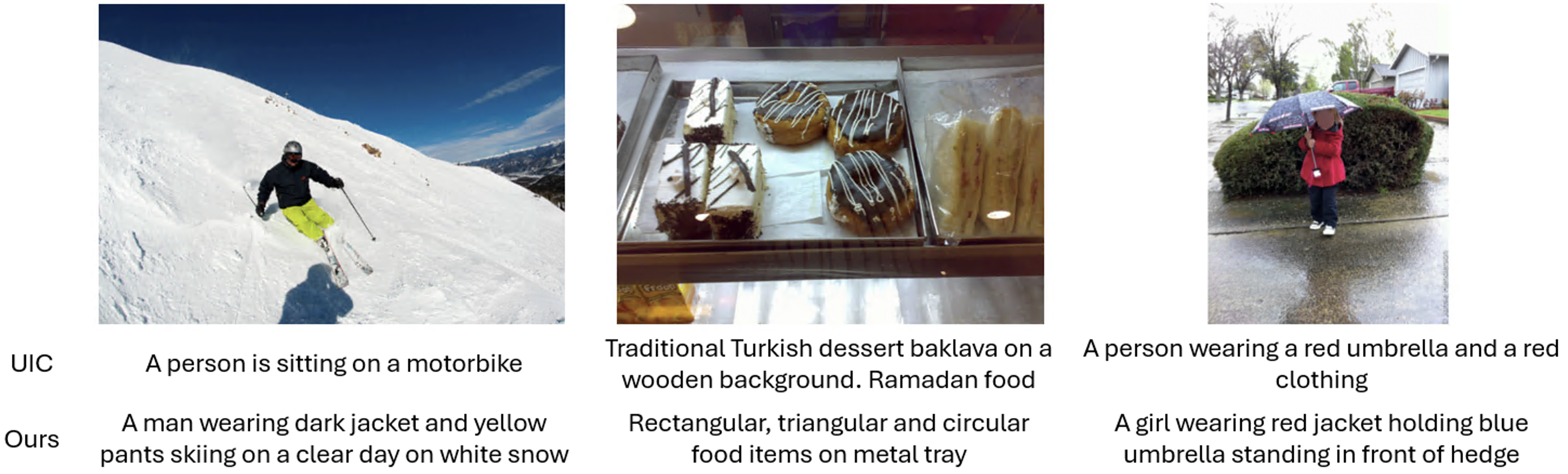}
    \caption{[Best viewed in color] Qualitative comparison of captions generated by UIC and our (unsupervised) method. In the central image, we can see that (rather than hallucinating) our method reverts back to the basics when it doesn't know the exact nature of the object in the given image and describes the shapes of the objects \ie, the preliminaries. 
    }
    \label{Figure: Qualitative Captions}
\end{figure*}

\paragraph{Training the \speaker}
The message generation procedure in the \speaker~is modeled as an RL problem. We consider an MDP with the rewards obtained from \methodName. A typical trajectory in this MDP will be as follows: $\tau=\{s_1, w_1, r_1, s_2, w_2, r_2, s_3,...,s_{T}, w_{T}, r_{T}\}$ where:
\begin{itemize}
    \item the $w_t$s are the words (or tokens) sampled from a vocabulary. They are analogous to the actions in the MDP and are sampled according to $w_t\sim\pi_1(s_t)$.
    \item the states are defined as $s_t=(s,w_1,...,w_{t-1})$, \ie the current state is the image received at the beginning along with the words predicted until the previous step.
    \item The $r_t$s are defined as $r_t=\gamma^{T-t}\mathcal{R}_1$, \ie the $r_t$s are actually the discounted rewards to go (rtgs).
\end{itemize}
The reward design merits more explanation. It can be noticed that discounting the rewards is a natural choice, as the entire sentence (if the correct description) should get a high reward, but the first word in the sentence should not get the same reward, since there are lots of chances to go wrong while sampling the subsequent words.

We train the decoder in $\mathcal{M}_1$ using the GRPO~\cite{shao2024deepseekmathpushinglimitsmathematical} PG algorithm variant where the loss function is:
\begin{align}
    \mathcal{L}_\mathrm{Speaker} = -\frac{1}{T}\sum_{t=1}^T\log\pi_1(w_t|s_t) (r_t-b)
\end{align}
where baseline $b=\sum_{t=1}^T{r_t}$ and the causal reward structure ensures low variance and helps in the convergence of the algorithm.

\paragraph{Training the \listener} The loss computation is relatively straight-forward for the \listener. Maximization of reward for the contextual MAB is equivalent to minimizing:
\begin{align}
    \mathcal{L}_\mathrm{Listener} = -\log\mathcal{R}_2
\end{align}
Since all we need are the probabilities to calculate the losses, we can pass the logits (outputs of $\pi_2(m)$) through a softmax layer to obtain the probabilities. Subsequently, we can back-propagate the losses through the decoder in $\mathcal{M}_2$ and update the \listener~weights.

The joint objective for learning the cooperative shared-reward strategy for playing \methodName~is:
\begin{align}
    \min \mathcal{L}_\mathrm{Speaker}+\lambda\mathcal{L}_\mathrm{Listener}
\end{align}
where $\lambda$ is a hyperparameter. It must be noted that the design of the aforementioned losses is necessary to enable distributed training across multiple GPUs, since, during implementation, there is no need for gradient copying across GPUs with the above loss objective design.

\section{Experiments}
\label{Section:Experiments}

In this section, we evaluate the effectiveness of our proposed method. To quantitatively evaluate our method, we use the images in the $2017$ version of the Microsoft COCO~\cite{10.1007/978-3-319-10602-1_48} dataset as the image set. MSCOCO has around $118$K labeled images and an additional $123$K unlabeled images. Following~\cite{feng2019unsupervisedimagecaptioning}, we use the $5$K MSCOCO validation split as the test set. We use an auxiliary dataset Flickr8k~\cite{flickr8k} with $8$K out-of-distribution images for initializing the cross-attention blocks in the \speaker~model (details follow). We also mix in the images from the Imagenet 21M dataset since our method doesn't require annotated images. The results are tabulated in~\Cref{Table: Main Results}.

\subsection{Implementation and Training}
\paragraph{Architecture Details} We use a larger Image Captioning model,  we use the Qwen-VL-2.5B version of the VLM in the speaker. For the smaller model, we use Vision Transformer (ViT) pre-trained on ImageNet $21$M for the image encoder and a modified version of \texttt{gpt-2-base}~\cite{radford2019language} in the \speaker. The ViT and GPT2 models were chosen because both of them have similar architectures with $12$ layers, each with $12$ attention heads, and $768$ dimensional hidden size, and are therefore compatible with each other. We couple the ViT and GPT by implementing a cross-attention block between them at every layer. All pre-trained weights are frozen, and only the cross-attention blocks and the LMHead layers are kept trainable. We initialize the weights of the \speaker~by using the Flickr8K dataset.

\begin{figure*}
    \centering
    \includegraphics[width=\linewidth]{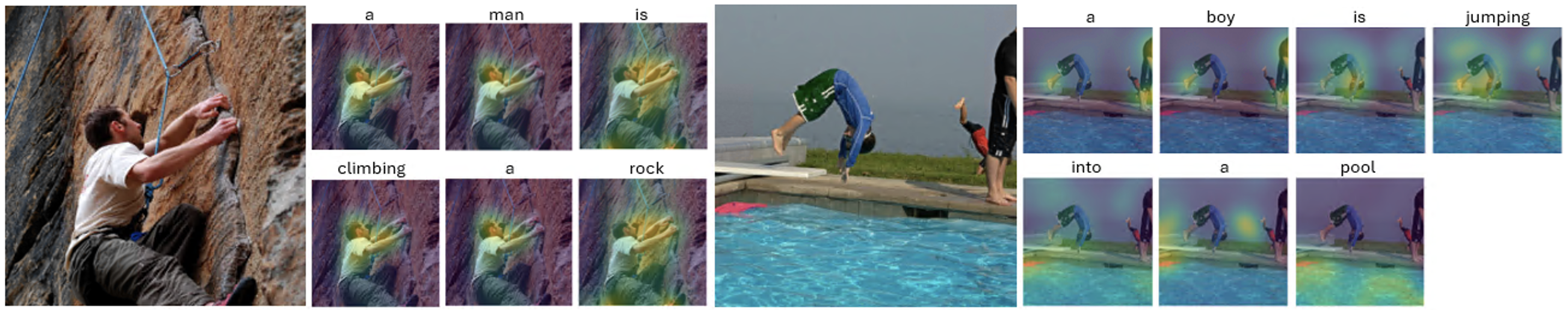}
    \caption{[Best viewed in color] A couple of predicted captions along with the heatmap over the image showing the region where the model focused its attention in order to predict each word.}
    \label{Figure: Attention}
\end{figure*}

As for the \listener, we use Microsoft Phi $2$~\cite{gunasekar2023textbooksneed} in the encoder, which is a $2.8$ billion parameter large language model. The vocabulary of the GPT is a subset of the vocabulary used by the \texttt{CodeGenTokenizer} of Microsoft Phi, hence we could directly send the tokens predicted by the \speaker~to the \listener~without further conversion. However, the Phi model uses $2560$ dimensional hidden embeddings. We use a $2$ layer MLP in \listener~with $2\times d_o$ neurons in the hidden dimension to convert the output embeddings of the LLM to the $d_o$ dimensional image embedding space where they can be compared to check similarity, for instance, $d_o=768$ for ViT. We use the same Vision encoder as in the \speaker~to get the image representations of the set of images shown to the \listener. We calculate the inner product to check the similarity of the embedding corresponding to the last token of the message with the image embedding corresponding to the \texttt{CLS} token of the ViT. Among all the image representations, the one with the maximum value of dot product is predicted by the \listener. It must be noted that our implementation of \listener~is independent of the number of distractors. A softmax function is applied over the inner products to get the probabilities.

\paragraph{Training Details} We use the PyTorch~\cite{pytorch} deep learning framework for running the experiments. We chose the contrastive set of images $K=1250$ for the experiments. We resize the images to $(h\times w)=(224\times224)$ before passing them through the ViT. We used the \texttt{timm} version of the ViT, hence we normalized by subtracting the mean $0.5$ and dividing by the standard deviation $0.5$. During training, we additionally first resize the images to $256\times256$ and randomly crop a patch $224\times224$ along with random horizontal flips. We used the vocabulary size $|\mathcal{V}|=50257$, keeping it the same as GPT2, and added one additional $768$ dimensional (trainable) parameter vector for the \texttt{<bos>} (beginning of a sentence) token. We allow the generated captions to be at most $32$ tokens long (excluding the special tokens). We fix $\lambda=1$ and $\gamma=0.95$ in all our experiments. We repeat our training $3$ times with random seed $2024$, $2025$, and $2026$. In order to scale up to $1250$ distractors, we distribute the training across $4\times$ NVIDIA A$100$ $40$GB GPUs. For the VLM \speaker, we use Low Rank Adaptation (LoRA) for parameter-efficient fine-tuning of the weights (we set the rank to $32$). We put $3$ versions of the \speaker~on $3$ GPUs and the \listener~on the remaining GPU. We used Adam optimizer with learning rates of $1e^{-5}$ for the trainable parameters in the \listener~and used SGD with learning rates $3e^{-4}$ for the trainable parameters in the \speaker models. We divide the images into 3 sub-batches and distribute them over the GPUs containing the \speaker~models. Once the captions are generated, we combine them and send them over to the remaining GPU for the \listener~model. Since the \speaker models get different batches of training data, after optimization, they have different weights. We sync the weights after every $5$ gradient step using NVLink. We use the Group Relative Policy Optimization variant of the Policy Gradient Reinforcement Learning Algorithm for improving backpropagating our rewards (we use $5$ generations for each image). We additionally clip the norm of the gradients to $1.0$ after back-propagation and before gradient update. On a node of $8\times \text{A}100$ GPUs, we are able to run 2 experiments, each lasting a couple of days, to finish training. The training codes, model checkpoints, and the wandb~\cite{wandb} training logs will be made public upon acceptance of the paper.

\subsection{Baselines}

We compare our methods with the following unsupervised baselines: \textbf{(i)} UIC~\cite{feng2019unsupervisedimagecaptioning}, \textbf{(ii)}~\citet{Laina_2019_ICCV}, \textbf{(iii)} IGGAN~\cite{10609416}, and \textbf{(iv)}~\citet{YANG2023102490}. This model performs unsupervised image captioning in remarkably similar manners by using an image set, a sentence corpus, and an existing visual concept detector. The sentence corpus is used to teach the captioning model how to generate plausible sentences, and the knowledge in the visual concept detector is distilled into the captioning model to guide the model to recognize the visual concepts in an image. The generated captions are kept semantically consistent with the image by projecting the image and caption into a common latent space so that they align with each other. We report the metrics of the aforementioned baselines verbatim from their corresponding papers.

\subsection{Metrics}
We use the following metrics that are standard in the unsupervised image captioning literature: \textbf{(i)} BLEU (Bilingual Evaluation Understudy), \textbf{(ii)} METEOR (Metric for Evaluation of Translation with Explicit Ordering), \textbf{(iii)} ROUGE (Recall-Oriented Understudy for Gisting Evaluation), \textbf{(iv)} CIDEr (Consensus-based Image Description Evaluation), and \textbf{(v)} SPICE (Semantic Propositional Image Captioning Evaluation).

\subsection{Ablation Experiments}
\label{SubSection:Ablation}

In order to understand the utility of each component in our method, we pose the following interesting research questions and design ablation experiments to answer the same:\\
\noindent\textbf{RQ1}: Why was an LLM used for language understanding in the \listener?\\
\noindent\textbf{RQ2}: What happens when we change the number of distractors?\\
\noindent\textbf{RQ3}: How well does the model adapt to novel objects?\\
\noindent\textbf{RQ4}: How sensitive is the procedure to the usage of pre-trained GPT for the caption generation part? Can we explain the captions generated?\\
\noindent\textbf{RQ5}: What happens if the model is initialized by supervised training parameters and then we make it play \methodName?\\
\noindent\textbf{RQ6}: Is the training algorithm stable? Or is it prone to mode collapse and over-fitting?

Consequently, we perform the following experiments:
\begin{enumerate}
    \item We replace the LLM with a \texttt{gpt-2-medium} $355$ million parameter model and train using our procedure.
    \item We set the number of distractors to $14$, $49$, $149$, $399$ and $1249$ and repeat the training procedure.
    \item We download a few images from the internet with unseen objects and perform inference on those images.
    \item We replace the \texttt{GPT-2-base} with a four layer LSTM~\cite{lstm} with Bahdanau Attention~\cite{bahdanau2016neuralmachinetranslationjointly} over the image embeddings and call this model ViT-LSTM \speaker. We fix the number of tokens to $3$K, the most frequently occurring words in the Flickr8K dataset, and initialize them with pre-trained GloVE~\cite{pennington-etal-2014-glove} weights. We used NLTK~\cite{nltk} to pre-process and tokenize the Flickr8K text corpus, and anything other than the top 3K words is replaced by the \texttt{<unk>} (unknown) token. We repeat our training procedure using the ViT-LSTM model.
    \item We train the ViT-LSTM model in the supervised setting and then retrain it to play \methodName~for around $12$ hours.
\end{enumerate}

\section{Results}
\label{Section:Results}

\Cref{Table: Main Results} shows that our proposed training approach obtains a clear advantage over existing unsupervised image captioning methods. We get better performance than the best-performing unsupervised baseline across all metrics. We also see that irrespective of the method used, \methodName can improve the image captioning performance, which is why we were able to improve the image captioning performance of encoder-decoder based VLMs using our \methodName. Additionally~\Cref{Figure: Qualitative Captions} shows some qualitative comparisons of captions generated by UIC and our method.

\Cref{Table: Ablation} shows that when the LLM in \listener~is replaced with a smaller \texttt{gpt} based language model, the BLEU score plummets. Surprisingly, though, we find that the game was solved $80\%$ of the time (a game is considered solved when the actual image is among the model's top 10 predictions during validation). We investigate the captions generated and find them to be gibberish from a human perspective. We realize that the \speaker~was able to find adversarial samples, and somehow both agents discovered a shared strategy for playing the game. Thus, we have answered \textbf{RQ1}.

\begin{table}[t]
    \centering
    \begin{tabular}{|l|c|}
        \hline
        Experiment & BLEU Score \\
        \hline
        Ours ($K=1250$, Ms Phi in \listener)  & 31.2 \\
        $K=400$  & 23.9 \\
        $K=150$  & 15.0 \\
        $K=50$  & 9.1 \\
        $K=15$  & 3.4 \\
        \listener~with \texttt{gpt-2-medium} & 0.6 \\
        ViT-LSTM \speaker~& 26.4 \\
        ViT-LSTM Supervised (w/o \methodName) & 32.8 \\
        ViT-LSTM Supervised + \methodName & 34.3 \\
        \hline
    \end{tabular}
    \caption{Reporting BLEU metrics for a wide variety of ablation experiments done to answer questions posed in~\Cref{SubSection:Ablation}.}
    \label{Table: Ablation}
\end{table}

For \textbf{RQ2}, we again refer to~\Cref{Table: Ablation}. We can see that as the value of $K$ decreases, the $BLEU$ score also decreases. We investigate the predicted captions from the learned model and find that the model has learned to only identify the colors in the background and/or the foreground. It so happens that with only, say, $14$ distractors, color identification of the subject is often good enough to reach the higher reward. The higher the value of $K$, the more precise the caption has to be in order for the \listener~to correctly identify the image shown to the \speaker.

\begin{figure}[t]
    \centering
    \includegraphics[width=\linewidth]{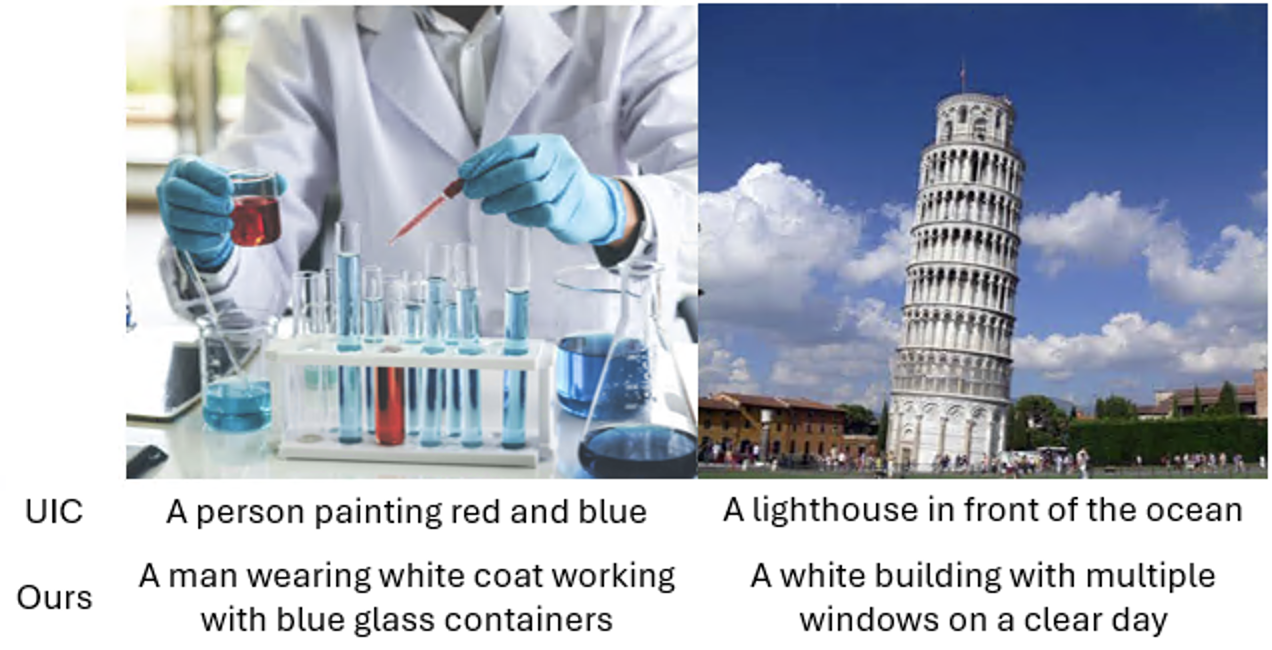}
    \caption{[Best viewed in color] Captions for a couple of images downloaded from the internet containing objects not in the training dataset of MSCOCO using our unsupervised method.}
    \label{Figure: OOD}
\end{figure}

\Cref{Figure: OOD} shows that our model is robust to novel objects in the image, thus answering \textbf{RQ3}. Since we do not use pre-trained object recognition models in our approach, it is less prone to hallucination.

To answer \textbf{RQ4}, we refer back to ~\Cref{Table: Ablation}. We see that due to the absence of pre-trained weights in the LSTM, we get slightly worse performance than GPT, but our method is certainly not restricted to GPT decoders. Additionally~\Cref{Figure: Attention} shows the outputs of the trained ViT-LSTM model for a couple of out-of-distribution images. We include the heatmap over regions where the LSTM was attending to while predicting each word.

\begin{figure*}
    \centering
    \includegraphics[width=\linewidth]{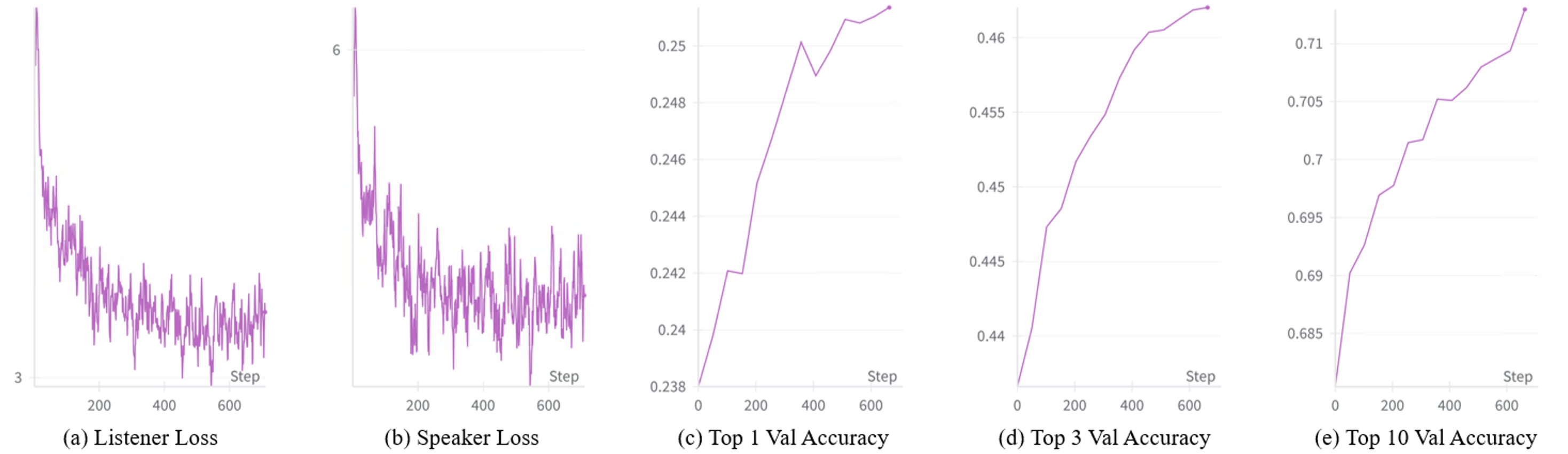}
    \caption{Training plots for ViT-LSTM with Bahdanau Attention after initializing with parameters learned using supervised learning and trained to play \methodName. We can see that the losses are stable, along with slight improvements in the validation accuracy, indicating that the model still has the capacity to learn, but the data was not adequate to train the model all the way.}
    \label{Figure: Plots}
\end{figure*}

~\Cref{Table: Ablation} additionally contains the performance metrics the ViT-LSTM model achieved when trained in a supervised setting. When we continued training in an unsupervised manner, we saw that we achieved higher BLEU scores than in the supervised setting. Thus, we have the answer to \textbf{RQ5}. We additionally plot the training losses and validation accuracies in~\Cref{Figure: Plots} of this particular set of experiments. \textbf{RQ6} is answered by observing that both the losses are going down simultaneously, and even though we are in a regime where usually models overfit and/or suffer from mode-collapse, our algorithm is very stable.

\section{Discussion}
\label{Section:Discussion}


\paragraph{MLE Formulation} Since the listener is given $K$ images to choose from, that \listener~objective can be written as
\begin{align*}
    \mathcal{L}_\mathrm{Listener} = -\sum_{k=1}^Ky_k\log\mathcal{R}_2=-\sum_{k=1}^Ky_k\log\pi_2(a=k|m)
\end{align*}
where $y$ is a one-hot vector with $1$ at the $k^\mathrm{th}$ location where the image is $s$. As we can see, this is nothing but the negative log-likelihood or the categorical cross-entropy loss. Thus, the \listener~is essentially trying to maximize the likelihood of it identifying the correct image among distractors.

Similarly, we can analyze the \speaker~objective. We can remove the baseline term as subtracting baseline is unbiased in expectation
is equivalent to:
\begin{align*}
    &\max \sum_{t=1}^T\log\pi_1(w_t|s_t) \gamma^{T-t} \sum_{k=1}^Ky_k\pi_2(a=k|m)\\
    \equiv &\max \sum_{t=1}^T\sum_{k=1}^K\gamma^{T-t}y_k\log\pi_1(w_t|s_t)\pi_2(a=k|m)\triangleq L
\end{align*}
Let $L_n$ be the expression for the $n^\mathrm{th}$ sample in the dataset and the combined parameters for the \speaker~and \listener~are $\theta$, the maximum likelihood (ML) estimate is:
\begin{align*}
    \theta^*_\mathrm{ML}=\argmax_\theta \frac{1}{N}\sum_{n=1}^N L_n
\end{align*}
which we can see is a joint maximization objective.

\paragraph{Relations to GANs}
One may think of the \speaker~as a GAN that generates natural language conditioned on images. It is possible to design a \listener~with the objective of detecting fake captions among a set of real captions. However, (i) it assumes access to captions, which would make the problem `unpaired' rather than `unsupervised', and (ii) it introduces a $\min \max$ adversarial objective over the parameters in \speaker~and Listener. Thus, this would be a saddle point optimization problem, which is notoriously difficult to train. Our problem, on the other hand, has a joint optimization objective, which is much more stable to train.

\section{Conclusion}
\label{Section:Conclusion}

We formulate unsupervised image captioning as a cooperative shared-reward strategic-form game. By scaling up the experiments, we learn a joint strategy to play the game and demonstrate both qualitatively and quantitatively that the strategy is superior to existing unsupervised approaches. We perform an extensive ablation study to determine the utility of each component in our proposed algorithm.

{
    \small
    \bibliographystyle{ieeenat_fullname}
    \bibliography{main}
}

\clearpage
\appendix
\setcounter{page}{1}
\maketitlesupplementary

\section{Dataset Details}

The MS COCO (Microsoft Common Objects in Context) dataset is a large-scale object detection, segmentation, key-point detection, and captioning dataset. The dataset consists of $328$K images.

Splits: The first version of the MS COCO dataset was released in $2014$. It contains $164$K images split into training ($83$K), validation ($41$K) and test ($41$K) sets. In $2015$, an additional test set of $81$K images was released, including all the previous test images and $40$K new images.

Based on community feedback, in $2017$, the training/validation split was changed from $83$K/$41$K to $118$K/$5$K. The new split uses the same images and annotations. The $2017$ test set is a subset of $41$K images of the $2015$ test set. The $2017$ release also contains a new unannotated dataset of $123$K images.

Annotations: The dataset has annotations for
\begin{itemize}
    \item object detection: bounding boxes and per-instance segmentation masks with $80$ object categories, captioning: natural language descriptions of the images (see MS COCO Captions).
    \item keypoints detection: containing more than $200$K images and $250$K person instances labeled with key points ($17$ possible key points, such as \texttt{left eye}, \texttt{nose}, \texttt{right hip}, \texttt{right ankle}),
    \item stuff image segmentation: per-pixel segmentation masks with $91$ stuff categories, such as \texttt{grass}, \texttt{wall}, \texttt{sky}.
    \item panoptic: full scene segmentation, For validation and testing, we use captions, \ie, only $10$K captions are used in total.
    \item dense pose: more than $39$K images and $56$K person instances labeled with DensePose annotations – each labeled person is annotated with an instance id and a mapping between image pixels that belong to that person's body and a template $3$D model. The annotations are publicly available only for training and validation images.
\end{itemize}

We only use the $236$K images for training, $5$K for validation, and additionally $5$K for testing. For validation and testing we use captions, \ie only $10$K captions are used in total.

\section{Execution on edge devices}
Using an off-policy Q-Learning algorithm, it is possible to perform sample training on edge devices using a 4-bit quantized version of the Phi model. This is useful since the captioning model can be initialized with pre-trained weights, and then, given enough photos on the device, the model can learn to describe the user images even when they are out of the distribution of the original training images. There are no privacy concerns in this case since the computations are on the device and can be run while the mobile device is charging during the night.

Additionally, the inference is very fast since generating captions does not use LLM, only the \speaker, which is architecture-agnostic. One can use something like LSTM decoders coupled with MobileNet for the image encoder to make edge device execution faster.



\section{Additional captioning examples}
\begin{table*}
    \centering
    \begin{tabular}{clcl}
        \multirow{2}{*}{\includegraphics[width=0.2\linewidth]{figures/3.jpg}} &
        \begin{tabular}[c]{@{}l@{}}\\ \\ \\ \\ \textbf{Supervised}: a man riding a\\skateboard on top of a ramp \\ \\ \end{tabular} & 
        \multirow{2}{*}{\includegraphics[width=0.2\linewidth]{figures/13.jpg}} & 
        \begin{tabular}[c]{@{}l@{}}\\ \\ \\ \\ \textbf{Supervised}: a room with a\\bed and a tv \\ \\ \end{tabular} \\
         & 
        \begin{tabular}[c]{@{}l@{}}\textbf{Supervised+}\methodName: a skate-\\boarder is doing a trick on a\\skateboard \end{tabular} &  & 
        \begin{tabular}[c]{@{}l@{}}\textbf{Supervised+}\methodName: a bed\\with a white blanket and a\\television \end{tabular} \\
        
        \multirow{2}{*}{\includegraphics[width=0.2\linewidth]{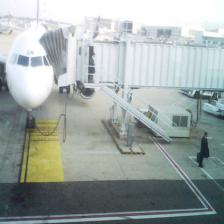}} & 
        \begin{tabular}[c]{@{}l@{}}\\ \\ \\ \\ \textbf{Supervised}: a plane is park-\\ed at the gate of an airport\\ \end{tabular} & 
        \multirow{2}{*}{\includegraphics[width=0.2\linewidth]{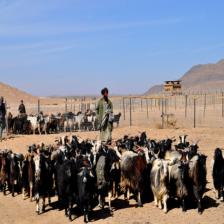}} & 
        \begin{tabular}[c]{@{}l@{}}\\ \\ \\ \\ \textbf{Supervised}: a man in a mili-\\tary uniform herding a flock\\of sheep \end{tabular} \\
         & 
        \begin{tabular}[c]{@{}l@{}}\textbf{Supervised+}\methodName: a white\\airplane parked at an airport\\loading passengers \\ \\ \end{tabular} &  & 
        \begin{tabular}[c]{@{}l@{}}\textbf{Supervised+}\methodName: a man\\herding a herd of goats in a\\field\end{tabular} \\
        
        \multirow{2}{*}{\includegraphics[width=0.2\linewidth]{figures/19.jpg}} & 
        \begin{tabular}[c]{@{}l@{}}\\ \\ \\ \\ \textbf{Supervised}: a street with a\\
        lot of cars parked in it \\ \\ \end{tabular} & 
        \multirow{2}{*}{\includegraphics[width=0.2\linewidth]{figures/25.jpg}} & 
        \begin{tabular}[c]{@{}l@{}}\\ \\ \\ \textbf{Supervised}: a sign that says\\$<\texttt{unk}>$ $<\texttt{unk}>$ $...$\\ $...$ $<\texttt{unk}>$\end{tabular} \\
         & 
        \begin{tabular}[c]{@{}l@{}}\textbf{Supervised+}\methodName: a bus\\is driving down the street in\\the middle of the street\end{tabular} &  & 
        \begin{tabular}[c]{@{}l@{}}\textbf{Supervised+}\methodName: a sign\\for a street sign on a city\\street\end{tabular} \\
        
        \multirow{2}{*}{\includegraphics[width=0.2\linewidth]{figures/27.jpg}} & 
        \begin{tabular}[c]{@{}l@{}}\\ \\ \\ \\ \textbf{Supervised}: a man sitting\\on a couch with a pillow on\\the back of his head \\ \\ \end{tabular} & 
        \multirow{2}{*}{\includegraphics[width=0.2\linewidth]{figures/28.jpg}} & 
        \begin{tabular}[c]{@{}l@{}}\\ \\ \\ \\ \textbf{Supervised}: a man and a\\little girl standing in front\\of a tall building \\ \\ \end{tabular} \\
         & 
        \begin{tabular}[c]{@{}l@{}}\textbf{Supervised+}\methodName: a wo-\\man sitting on a couch in a\\train seat\end{tabular} &  &
        \begin{tabular}[c]{@{}l@{}}\textbf{Supervised+}\methodName: a man\\is sitting on the concrete\\with a bird\end{tabular} \\
        
        \multirow{2}{*}{\includegraphics[width=0.2\linewidth]{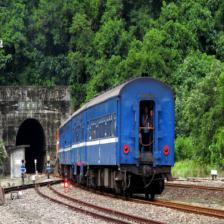}} & 
        \begin{tabular}[c]{@{}l@{}}\\ \\ \\ \\ \textbf{Supervised}: a train is travel-\\ing down the tracks in the day \\ \\ \end{tabular} & 
        \multirow{2}{*}{\includegraphics[width=0.2\linewidth]{figures/63.jpg}} & 
        \begin{tabular}[c]{@{}l@{}}\\ \\ \\ \textbf{Supervised}: a man riding skis\\down a snow covered slope\end{tabular} \\
         & 
        \begin{tabular}[c]{@{}l@{}}\textbf{Supervised+}\methodName: a blue\\and white train traveling\\down tracks \\ \\ \end{tabular} &  &
        \begin{tabular}[c]{@{}l@{}}\textbf{Supervised+}\methodName: a man\\on skis standing in the snow\end{tabular} \\
    \end{tabular}
    \caption{[Best viewed in color] Qualitative comparison of ViT-LSTM model after supervised training and the same model after additional unsupervised training using \methodName. It can be observed that after \methodName~training, captions are more detailed and less erroneous. \methodName includes color attributes in captions (top right, bottom left). In the middle left image, \methodName~inferred a bus just from a partial image of the same (the rear-view mirror). However, \methodName~has no counting mechanism (last but one on the right).}
    \label{Table:Appendix}
\end{table*}

\end{document}